\documentclass[]{article}
\usepackage{algorithm}
\usepackage[noend]{algpseudocode}
\usepackage{algorithmicx}
\usepackage{multirow}
\usepackage{graphicx}
\usepackage{subcaption}
\usepackage{amsmath}
\usepackage{amsfonts}
\usepackage[a4paper, total={6in, 8in}]{geometry}
\usepackage[inline]{enumitem}
\usepackage{hyperref}

\makeatletter
\def\blfootnote{\gdef\@thefnmark{}\@footnotetext}
\makeatother

\usepackage{natbib}
\bibpunct[, ]{(}{)}{,}{a}{}{,}%

\algnewcommand\ReturnNNL{\State \algorithmicreturn{} }
\newcommand{\stt}{\; | \;}
\def\argmin{\mathop{\rm arg\,min}}%

\author{
  Andr\'{e} Hottung\\
  Bielefeld University\\
  \texttt{andre.hottung@uni-bielefeld.de}
  \and
  Shunji Tanaka\\
  Kyoto University\\
  \texttt{tanaka@kuee.kyoto-u.ac.jp}
  \and
  Kevin Tierney\\
  Bielefeld University\\
  \texttt{kevin.tierney@uni-bielefeld.de}
}
\title{Deep Learning Assisted Heuristic Tree Search for the Container Pre-marshalling Problem}
\date{\vspace{-2ex}}

\newcommand{\add}[1]{#1}

\begin{document}

\maketitle

The container pre-marshalling problem (CPMP) is concerned with the re-ordering of containers in container terminals during off-peak times so that containers can be quickly retrieved when the port is busy. The problem has received significant attention in the literature and is addressed by a large number of exact and heuristic methods. Existing methods for the CPMP heavily rely on problem-specific components (e.g., proven lower bounds) that need to be developed by domain experts with knowledge of optimization techniques and a deep understanding of the problem at hand. With the goal to automate the \add{costly} and time-intensive design of heuristics for the CPMP, we propose a new method called Deep Learning Heuristic Tree Search (DLTS). It uses deep neural networks to learn solution strategies and lower bounds customized to the CPMP solely through analyzing existing (near-) optimal solutions to CPMP instances. The networks are then integrated into a tree search procedure to decide which branch to choose next and to prune the search tree. DLTS produces the highest quality heuristic solutions to the CPMP to date with gaps to optimality below 2\% on real-world sized instances.

\blfootnote{\textcopyright 2019. This manuscript version is made available under the CC-BY-NC-ND 4.0 license \url{http://creativecommons.org/licenses/by-nc-nd/4.0/}. The formal publication of this manuscript is available at \url{https://doi.org/10.1016/j.cor.2019.104781}.}

\section{Introduction}
The throughput of containers at the world's seaports has been growing at a tremendous rate. From 2010 to 2017 the amount of containers shipped increased by 34\% from 560 to 753 million twenty-foot equivalent units (TEU) \citep{UNCTADstat}. The rising volume poses a major challenge for port operators, who must quickly transfer millions of containers between modes of transportation \citep{RoCoSl09}. Frequent delays at a port lead to shippers shifting to more reliable locations, resulting in a loss of business. It is therefore of great interest for port operators to prevent delays.  

Delays can occur at various points at a port, including the transfer of containers between terminals (inter-terminal transportation) or in intra-terminal operations. We address the latter, focusing on delays caused when storing and retrieving containers in the yard. Two key problems arise in this context: The container relocation problem (CRP) and the container pre-marshalling problem (CPMP). We provide a new solution procedure for the CPMP, which is a housekeeping problem first introduced by~\cite{LeCh09} in which a rail-mounted gantry crane is used to re-order containers during off peak times so that they can be quickly extracted when the port is busy. The goal of the problem is to find a minimal sequence of container movements that sort a set of container stacks according to the time each container is expected to exit the yard.

A number of methods have been developed to solve the CPMP, including several \add{exact} approaches (e.g.,~\cite{lee:hsu:07}, \cite{rendlprand}, \cite{BrZw14}, \cite{tanaka2017}). However, all these approaches still need too much time to solve real-world sized instances to be used in a decision support system. Thus, a large number of heuristic methods for the CPMP have been proposed (e.g.,~\cite{LeCh09}, \cite{caserta:voss:09Coll}, \cite{Exposito}). All of these methods rely heavily on problem-specific components \add{(e.g., proven lower bounds or local search procedures)} painstakingly developed by domain experts. Developing these components requires not only knowledge of optimization techniques but also a deep understanding of the CPMP.

We develop a new method that we call Deep Learning Heuristic Tree Search (DLTS) with the goal of enabling the automated generation of heuristics for the CPMP. We integrate deep (artificial) neural networks (DNNs) into a heuristic tree search (HTS) to decide which branch to choose next (branching) and to estimate a bound for pruning the search tree (bounding). The DNNs are trained offline via supervised learning on existing (near-) optimal solutions for the CPMP and are then used to make branching and bounding decisions during the search. 

DLTS contains a number of configurable components. Search strategies and decisions as to how to use information from the DNNs can be configured offline using an algorithm configurator, such as GGA~\citep{gga} or GGA++~\citep{gga++}, to increase the quality of the solutions found.

DLTS is able to achieve a high level of performance \add{although no CPMP specific heuristics are explicitly encoded in the search procedure}; problem specific information is almost exclusively provided as input to the DNN. After DLTS has been trained on existing solutions, it can be used to find high quality heuristic solutions
in a fraction of the run time of an exact approach. We show experimentally how DLTS is also able to significantly outperform the state-of-the-art heuristic approaches, finding gaps to optimality between 0\% and 2\% on real-world sized instances compared to gaps of 6\% - 15\% for state-of-the-art metaheuristics.

The main contributions of this paper can be summarized as follows: 
\begin{enumerate}
    \item The first tree search algorithm for an optimization problem with branching and bounding decisions made entirely through a learned model.
    \item The highest quality heuristic solutions for the CPMP to date. 
    \item An experimental evaluation of different search strategies and DNN architectures for DLTS.  
\end{enumerate}

This paper is organized as follows. First, we discuss related work for the CPMP and the area of combining \add{machine learning} and optimization techniques in Section~\ref{sec:relwork}. We then introduce the CPMP and the DLTS algorithm along with several search strategies and parameterizations in Section~\ref{sec:cpmp} and Section~\ref{sec:dlts} accordingly. This is followed by a description of the application of DLTS to the CPMP in Section~\ref{sec:cpmp_dlts}. In Section~\ref{sec:compres} we test our approach experimentally on a large dataset of CPMP instances. We conclude and discuss future work in Section~\ref{sec:concl}.

\section{Related work}
\label{sec:relwork}
In this section we first provide an overview of existing methods for the CPMP. We then continue with a discussion of methods that combine learning and optimization similar to DLTS. DLTS is the first algorithm that uses DNNs to guide a tree search for an optimization problem. However, a (quickly growing) number of optimization approaches already exist that integrate machine learning methods in other areas. We provide an extensive overview of these approaches (many of which have inspired parts of DLTS) and discuss them in relation to DLTS.

\subsection{Container Pre-Marshalling Problem}
Since the introduction of the CPMP by~\cite{LeCh09} a large number of \add{exact} and heuristic methods have been proposed. \add{Exact} methods include the integer programming approach in~\cite{lee:hsu:07}, the constraint programming model in~\cite{rendlprand}, the branch-and-price algorithm in~\cite{BrZw14}, an A*/IDA* technique in~\cite{idacpmp}, and an iterative deepening branch-and-bound algorithm in~\cite{tanaka2017}.

Heuristic approaches focus on generating solutions quickly and allow a real world application even in the case of bays with a large number of stacks and tiers. \cite{caserta:voss:09Coll} introduce the corridor method\add{, an algorithm that} creates a so-called ``corridor'' within the bay to limit the number of possible moves. Additionally they use a local search procedure that moves containers according \add{to} a set of predefined rules. The lowest priority first heuristic (LPFH) proposed by~\cite{Exposito} tries to move containers with a ``low priority'' (i.e., late exit time) first. This method outperforms the corridor method, especially on smaller instances. In~\cite{JovanovicVoss:13}\add{,} LPFH is extended with a multistart strategy and a complex set of problem-specific rules to choose \add{where each container should be relocated (e.g., a look-ahead method)}. \cite{bortfeldt2012} introduce a novel lower bound and a heuristic tree search using a branching schema with move sequences instead of single moves. They report an improved performance in comparison to  \cite{caserta:voss:09Coll}.  \cite{WaJiLi15} propose a target guided approach within a beam search, and in~\cite{brkgacpmp} a biased random-key genetic algorithm (BRKGA) with a decoder to construct a solution is used. Especially on larger instances both methods significantly outperform the tree search approach from \cite{bortfeldt2012}, with the BRKGA needing less than a minute for the solution generation.

The \add{CRP (also known as the block(s) relocation problem)} is closely related to the CPMP and has been thoroughly investigated in the literature. In contrast to the CPMP, the CRP tries to reduce the number of container movements when retrieving containers from a bay, meaning in each step of solving the problem a container is removed from the bay. We focus on two recently proposed approaches, because of their similarity to DLTS. \cite{ku2016abstraction} propose a\add{n} abstraction method to reduce the search space of the CRP together with an offline generated pattern \add{database} that stores optimal solutions for abstract states at a certain level of the search tree. \cite{quispe2018exact} use a pattern database in a similar manner in a exact iterative deepening A* procedure together with two new proposed lower bounds. Similar to DLTS, the approaches rely on existing solutions generated in an offline phase that only amortizes if many instances of the problem have to be solved. However, they only store these solutions in a pattern database (i.e., a lookup table) and do not use any learning to generalize beyond seen states. Furthermore, the pattern database is only used at a predefined level in the search tree, whereas in DLTS \add{the branching decisions are made with a learned model} at all levels of the search tree.

\subsection{Machine Learning and Optimization}
Learning mechanisms have been successfully applied within search procedures to select which heuristics to apply online (e.g. the DASH method introduced by~\cite{liberto2016}. Furthermore, in algorithm selection techniques a machine learning model is used to attempt to choose the best algorithm out of a portfolio of options for a given problem instance. These methods have been applied to a number of problems. See, e.g.,~\cite{aslib} \add{and~\cite{kotthoff2016algorithm}} for an overview. \add{Our work is inspired by the approach of~\cite{silver2016}, in which two DNNs are used to guide a Monte Carlo tree search to play the game Go.}

We split our further discussion of literature regarding learning and optimization into \add{three} parts. First, we describe  approaches that use machine learning techniques to obtain an exact solution. We then make note of relevant combinations of learning techniques in heuristics. \add{Finally, we discuss approaches using deep learning to solve optimization problems.}





\subsubsection{Learning in exact solvers}

\add{\cite{lodi2017}, together with comments from~\cite{dilkina2017}, provide an overview of methods applying learning to the problems of variable and node selection in \add{mixed-integer programming (MIP)}. Some of the articles identified by~\cite{lodi2017} are of particular relevance to our work on DLTS so we describe them here.}


Several methods have been developed to provide a surrogate for \emph{strong branching} scores, which are a way of ranking the possible branches during a MIP branch-and-bound search.
These approaches approximate the scores faster than the true values can be calculated. \cite{khalil2016} learn a model for predicting the \add{ranking} of the scores of strong branching. They use features derived from the search trajectory and show speed\add{-}ups using their method versus CPLEX. \cite{alvarez2017} also approximate strong branching scores.
In contrast to these methods DLTS is trained on (near-) optimal solutions rather than strong branching scores or other values produced during search.
Furthermore, the predictions from the DLTS branching network form more than a simple ranking over branching decisions. The branching network produces a probability distribution over branches, which provides a confidence level in each branch.

A logistic regression is used in~\cite{khalil2017} to predict when to apply a primal heuristic when solving a MIP. The authors use similar features to~\cite{khalil2016} and are able to improve the performance of a MIP solver. Other approaches using machine learning techniques to solve a MIP are proposed in~\cite{kruber2017},
~\cite{bonfietti2015}, and~\cite{lombardi2017}.

\cite{vaclavik2018} improve the performance of a branch-and-price algorithm by predicting an upper bound for each iteration of the pricing problem using online machine learning. They evaluate their method on the nurse rostering problem and on a scheduling problem, observing a 40\% and 22\% CPU time reduction on average, \add{respectively}.

\add{We note that in contrast to the methods discussed in this section, DLTS does not make branching decisions in an exact branch-and-bound search (e.g., in a MIP). Instead, DLTS searches the space of sequences of container movements with branching decisions determining the sequential construction of CPMP solutions.}

\subsubsection{Learning (in) heuristics}

To the best of our knowledge, the first proposed use of learning methods within a heuristic search procedure comes from Glover's \emph{target analysis} technique \citep{glover1989,glover1986}. The idea is to rate each branch based on a weighted sum of criteria and choose the branch with the highest rating. The weights can be adjusted offline using a learning procedure. A recent realization of this technique is \emph{hyper configurable reactive search}, introduced in~\cite{ansotegui2017}, in which the parameters of a metaheuristic are determined online with a linear regression. The weights of the regression are tuned offline with the GGA++ algorithm configurator~\citep{gga++}.

Algorithms for ``learning to search'' used to solve structured prediction problems also perform a heuristic search. To this end the structured prediction problem is first converted into a sequential decision making problem, for which a policy is then learned/improved. Learning to search methods include SEARN~\citep{daume2009} and LOLS~\citep{chang2015}.   
A key limitation of these approaches is that they use a greedy search at test time, meaning that there is no mechanism for correcting ``mistakes'' (deviations from the optimal solution sequence). 

\add{\cite{he2014} propose a method to learn a node ordering over open nodes in a heuristic branch-and-bound search using imitation learning. They categorize their features into three groups: node features (e.g., lower bound, depth), branching features (e.g., pseudocost) and tree features (e.g., global upper/lower bounds).  The features are similar to those used in the previously mentioned DASH approach, which is exact and involves a branch-and-bound search. In DASH, the features try to represent the characteristics of the remaining subproblem (e.g., percentage of variables in the subproblem; depth in the tree). In contrast to DASH, the method of \cite{he2014} identifies the next node to explore during search instead of selecting a branching heuristic at a node.}

\cite{karapetyan2017} propose a metaheuristic schema that allows for the automated generation of multi-component metaheuristics by learning transition probabilities between single heuristic components (being either hill climbing or  mutation operators). The approach is flexible enough to model several standard metaheuristics, and the best learned metaheuristic for the bipartite boolean quadratic programming problem is significantly faster than previous methods.


\subsubsection{\add{Deep learning for optimization problems}}


\add{\cite{vinyals2015pointer} introduce a so-called \textit{pointer network} (a special type of DNN) and train it to output solutions for the traveling salesman problem using supervised learning. \cite{bello2016neural} train a pointer network for the traveling salesman with reinforcement learning. \cite{kool2018attention} propose a similar approach that can also be used to solve other routing problems, such as the vehicle routing problem.  \cite{dai2017} train a graph embedding network with reinforcement learning to generate solutions for graph problems (e.g., minimum vertex cover and maximum cut problem). All approaches focus on the training and the architecture of the DNNs instead of how the DNNs can be incorporated into a sophisticated search procedure. Even though the results are promising, the approaches can not compete with state of the art approaches on larger instances.}

\add{Recently, DNNs have also been used in the context of the constraint satisfaction problems (CSPs). \cite{xu2018towards} successfully use a convolutional DNN to predict the satisfiability of random Boolean binary CSPs. \cite{galassi2018model} probe if a DNN can learn to construct a solution for a CSP by training it to make a single variable assignment using supervised learning techniques.}

\section{Container Pre-Marshalling}
\label{sec:cpmp}

\begin{figure}
    \begin{center}
        \begin{subfigure}[b]{0.4\textwidth}
            \includegraphics[width=\textwidth]{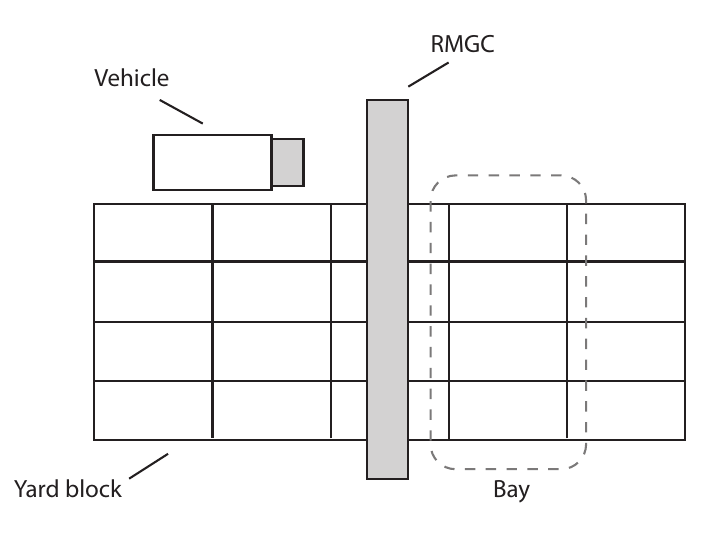}
            \caption{Top view} 
        \end{subfigure} \;
        \begin{subfigure}[b]{0.4\textwidth}
            \includegraphics[width=\textwidth]{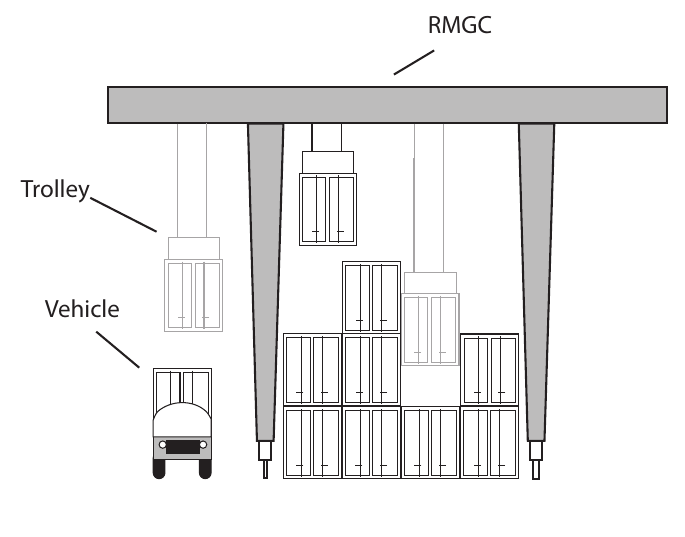}
            \caption{Front view}
        \end{subfigure} \;
    \end{center}
    \caption{A yard block with a rail-mounted gantry crane (RMGC), from~\cite{idacpmp}}
    \label{fig:bay}
\end{figure}

In a container terminal, containers are stored in a large buffer area, called the yard, while they wait to be transferred to a ship or to other modes of transportation. The CPMP is concerned mainly with yards in which rail mounted gantry cranes are used to store and retrieve containers. The containers are usually organized into rectangular blocks containing multiple rows of container stacks. A single row of stacks forms a bay (shown in Figure~\ref{fig:bay}). All stacks of a bay have a common height restriction (usually due to the height of the crane) measured in tiers of containers.

The CPMP arises when containers stacked in a single bay need to be re-sorted so that they can be quickly extracted. Each container is assigned a \emph{group} that corresponds to the scheduled exit time of the container from the bay. If a container with a late exit time is stacked on top of a container with an early exit time, it blocks the removal of that container and must be re-stowed during port operations, wasting valuable time. Only a single crane is available to move one container at a time from the top of one stack to the top of another stack. The idea of pre-marshalling is to re-sort the containers with a minimum number of container movements during off-peak times, so that container retrieval operations run smoothly when the port is busy. The CPMP is NP-hard \add{\citep{BrZw14}}.

\subsection{Formal problem definition}

The CPMP involves a set of $C$ containers arranged into $S$ stacks that have a maximum height $T$. The parameter \add{$g_{ij}$} provides the group value (retrieval time) of the container in stack \add{$i$} at \emph{tier} (height) \add{$j$}. The objective of the CPMP is to find a minimal length sequence of \add{stack-to-stack} movements \add{$(i,i')$} in which a container is moved from the top of stack \add{$i$} to the top of stack \add{$i'$}, such that all stacks are sorted, i.e., \add{$g_{ij} \geq g_{i,j+1}, \forall \; 1 \leq i \leq S, 1 \leq j < T$}.

\begin{figure}
    \centering
    \includegraphics[width=0.8\textwidth]{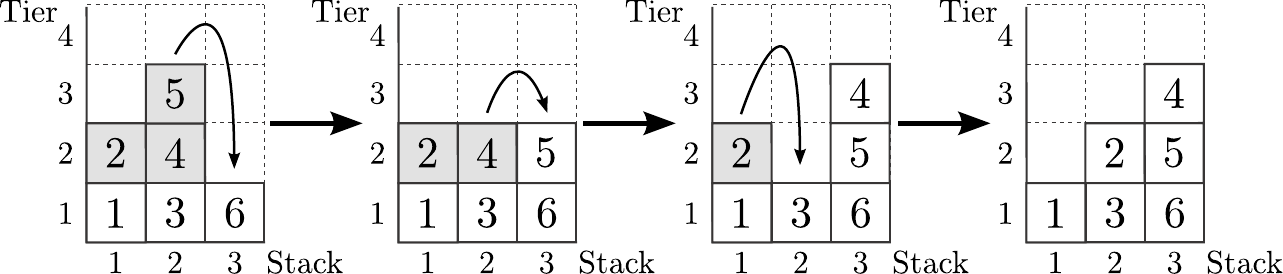}
    \caption{Example CPMP solution from~\cite{brkgacpmp}.}
    \label{fig:pmex}
\end{figure}

Figure~\ref{fig:pmex} shows a CPMP problem instance and its optimal solution. Starting on the left, there are three stacks with a total of six containers, each one labeled with its group (note that multiple containers can have the same group value, but for ease of presentation, we assign a unique group to each container). The containers in gray are in blocking positions and must be moved so that they are not blocking any containers beneath them. \add{Notice how the search for a solution to the CPMP can be naturally mapped to a search tree with the nodes of the tree representing the configuration of containers in the bay, and the branches between the nodes representing the possible movements.}

\add{Existing methods often use lower bounds to prune the search space. The simplest lower bound is given by counting the number of blocking containers (e.g., at least three movements are necessary to sort the stacks of the instance shown in Figure~\ref{fig:pmex}).
Improved lower bounds have been introduced, among others, by~\cite{bortfeldt2012},~\cite{tanaka2017} and~\cite{tanaka2019branch}. These bounds are computationally efficient to compute, but often have multiple move gaps to the optimal solution value.  
In DLTS we do not use any of the lower bounds from the literature; instead we use a DNN to heuristically determine the lower bounds.
}

\section{Deep Learning assisted heuristic Tree Search}
\label{sec:dlts}

\newcommand{\sdepth}{\textsc{Depth}}
\newcommand{\mprob}{\textsc{MP}}
\newcommand{\ssucc}{\textsc{Successors}}
\newcommand{\scost}{\textsc{Cost}}
\newcommand{\scomplete}{\textsc{Complete}}
\newcommand{\ub}{\mathit{ub}}
\newcommand{\lb}{\mathit{hlb}}
\newcommand{\costc}{\mathit{c}_{cur}}
\newcommand{\costb}{\mathit{c}_{best}}
\newcommand{\best}{$n$*}
\newcommand{\bestm}{\add{n}$*$}
\newcommand{\bestml}{$M$*}
\newcommand{\bestmlm}{M$*$}
\newcommand{\dfsstack}{T}
\newcommand{\dnnp}{\textsc{DNN-Branching}}
\newcommand{\dnnv}{\textsc{DNN-Bounding}}
\newcommand{\branchvals}{\textit{branch-values}}

\add{DLTS consists of a heuristic tree search in which decisions about which branches to explore and how to bound nodes are made by DNNs. Each time a new node is opened in the search tree, the so called \emph{branching DNN} is used to decide
\begin{enumerate*}[label=\textit{\alph*)}]
\item which branches should be pruned and
\item in which order the non-pruned child nodes should be explored.
\end{enumerate*}
The order in which nodes are visited throughout the search is determined by traditional strategies like depth first search (DFS), limited discrepancy search (LDS)~\citep{harvey1995}, and weighted beam search (WBS). Additionally, we also use a so called \emph{bounding DNN} at some levels of the tree to determine a lower bound that prunes nodes, thus reducing the branching factor.}

\add{In this section, we first describe how optimization problems can be solved using tree search based methods. We then explain in detail how DLTS uses DNNs to make branching decisions and to compute lower bounds during the search. Finally, we present the different search strategies for DLTS and describe three possible ways to prune the branches of a node based on the output of the branching network.}

 \subsection{\add{Tree Search}}
\add{Tree search based methods are frequently used to solve optimization problems. Starting at the root node, the search tree is explored by systematically exploring the child nodes of the root node and subsequent nodes.}
A \add{complete} solution to a given optimization problem \add{can be understood as} a path from the root node to a feasible leaf node, consisting of a sequence of \add{$m$} branching decisions \add{$b_1, \dots, b_m$}. The state $s_0$ is the initial state represented by the root node.
The objective value (hereafter referred to as the cost) of a complete solution is denoted by \add{$c$*}. The cost of a partial solution \add{$b_1, \dots, b_k$} is denoted \add{$c_k$}.

In the case of the CPMP each node in the tree represents a container configuration with the initial container configuration being represented by the root node. The child nodes of a node represent all possible container configurations that can be reached by one container movement. A \add{complete} solution for the CPMP is then given by the path from the root node to a node representing a sorted container configuration. \add{The cost associated with a solution $b_1, \dots, b_k$ for the CPMP is $k$ (i.e., each container movement increases the cost by 1).}
In the next section we describe how DNNs can be used to focus the search on promising areas of the tree and to provide lower bounds throughout the search.

\subsection{DNNs for Tree Search}

DNNs are function approximators inspired by biological neural networks. A DNN consists of multiple layers of \emph{perceptrons} (\add{neurons}). Each \add{neuron} accepts one or more weighted inputs from \add{neurons} of the previous layer, aggregates those inputs, and applies an \emph{activation function} to the inputs. The value from this function is then sent out to the \add{neurons} of the next layer. The DNN ``learns'' by \add{optimizing} the weights on the arcs of the network. In this work, we use DNNs purely in a supervised fashion. For more detail regarding DNNs we refer to~\cite{goodfellow2016}.

There are three main types of layers for a DNN: the input layer that accepts \add{an input} and transmits it into the network; an output layer that consolidates the information of the network into a set of outputs; and hidden layers, which accept and re-transmit data through the network. The layers are organized sequentially, starting with an input layer, followed by one or more hidden layers, ending with the output layer.

Consider a standard supervised learning setting in which the goal is to learn a function $f : X \rightarrow Y$, where $X$ is the input space and $Y$ is the output space. DNNs can be used for both classification (the space $Y$ consists of a set of discrete values) as well as regression ($Y$ can take any value in $\mathbb{R}$), and we use both types of DNNs in this work. \add{We use the branching network to make predictions about which branch will be best (classification DNN) and the bounding network to predict the cost of completing a solution for a node in the search tree (regression DNN)}.

\begin{figure}
    \centering
    \includegraphics[width=0.85\textwidth]{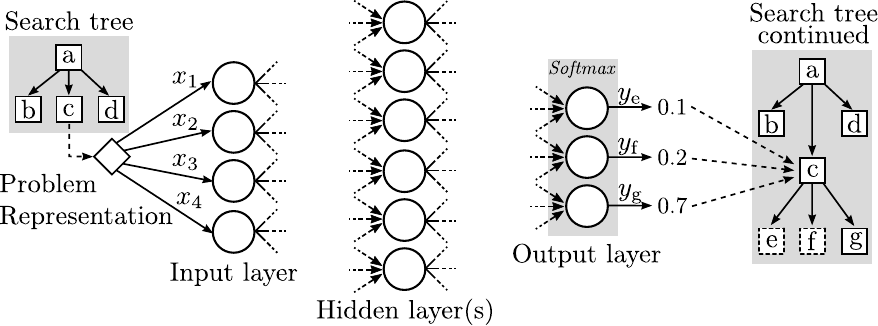}
    \caption{\add{Overview of a branching decision for DLTS.}}
    \label{fig:dnn}
\end{figure}

\add{The branching DNN in DLTS is used as follows.} When a node \add{$n_k$} is reached in the search the associated state \add{$s_k$} is given to the network. The input is then propagated through the branching network, which has as many outputs as there are possible branches for node \add{$n_k$}. \add{For the CPMP the number of branches depends only on the number of stacks of an instance and is thus the same for all nodes.} We use a \emph{softmax} activation function in the output layer to transform all of the outputs into values in $[0,1]$ such that they sum to~1. This allows DLTS to use the output as a probability distribution over the available branches. The output is then used to decide which branches of node \add{$n_k$} should be explored (e.g. exploring the branch associated with the highest output first). We note that this distribution provides significantly more information than just a ranking, as the probability assigned to a branch indicates the DNN's confidence in \add{this branch. Branches assigned low probability values by the network} can, for example, be discarded. \add{Figure~\ref{fig:dnn} shows an example of the branching within DLTS. In this case, the nodes $e$ and $f$ are not explored because of the low probability of leading to an optimal solution, as assigned by the branching DNN.}

The bounding network has a similar architecture except for the output layer, \add{which only consists of a single output.} It is given the same input as the branching network (state \add{$s_k$} associated with a node \add{$n_k$}) and predicts the cost of completing the associated solution. \add{The heuristic lower bound is then given by} the cost \add{$c_k$} of the partial solution \add{$b_1, \dots, b_k$} associated with node \add{$n_k$} plus the predicted cost of the \add {bounding} network to complete \add{$b_1, \dots, b_k$.} \add{If the heuristic lower bound} exceeds or is equal to the cost of the current best solution, no branches of node \add{$n_k$} should be explored. Because the prediction of the bounding network is subject to errors, it can \add{be} multiplied by a factor between 0 and 1 to reduce \add{the heuristic lower bound}.

\subsubsection{DNN Training}
\label{sec:training}

Training for the DLTS branching and bounding networks works as follows. A set of representative instances \add{is} split into a training and a validation set. The instances are solved using an exact procedure, although a heuristic could be used if no exact algorithm \add{were} available. A DNN training set is then created by examining each optimal (or near\add{-}optimal) solution and extracting DNN training examples.

\medskip

\add{For the CPMP, a} complete solution is a sequence of \add{$m$} movements \add{$(i_1,i'_1),...,(i_m,i'_m)$} in which in step \add{$k$} a container is moved from the top of stack \add{$i_k$} to the top of stack \add{$i'_k$} (with \add{$i_k \neq i'_k$}). Let \add{$B_k$} be \add{a matrix representing} the state of the instance before move \add{$k$} is performed, where \add{$B_{kij}$ is the group value of the container in stack $i$ at tier $j$ before move $k$}. Empty positions are assigned the value zero. The output space of the DNN is the space of all possible moves $\{1,\dots,S\} \times \{1,\dots,S\} \setminus \{(1,1),  (2,2),\dots,(S,S)\}$. \add{Thus, the output space only depends on the number of stacks $S$ of an instance. Infeasible moves (e.g., moving a container to a already full stack) are filtered out in a subsequent step (see Section~\ref{sec:knowledge}).}

For each container movement \add{$k$}, we create a training example \add{$(x_k, y_k)$ for the branching network, with $x_k \in X$ and $y_k \in Y$}, where \add{$x_k := B_k$ and $y_k := \Delta_k$.} We let \add{$\Delta_k$ be} a vector of $S(S-1)$ entries with 
\begin{equation}\Delta_{kss'} := 
\begin{cases} 
    1 & \text{ if } s = i_k \wedge s' = i'_k \\
    0 & \text{ otherwise. }
\end{cases}
\label{kss}
\end{equation}
This provides both positive and negative information about what branches lead to an optimal solution to the \add{branching} DNN. We note, however, that \add{while we currently only consider one optimal solution per instance,} other training schemes could be possible, such as when multiple optimal solutions are available for a particular instance, \add{which is often the case for the CPMP}.

For training the \add{bounding} network, we use similar input as for the branching network. The key difference is that instead of an output for each branch, the value network has a single output that provides an estimate of the cost to complete a partial solution. We thus create training examples with \add{$x_k := B_k$} and \add{$y_k := m - k + 1$}.

\medskip

\add{During training, each DNN is repeatedly presented with a small sample $\{(x_{q_1},y_{q_1}),\dots,(x_{q_v},y_{q_v})\}$ of the DNN training set. The input $x_{q_1},\dots,x_{q_v}$ is} propagated through the network to generate the associated output \add{$f(x_{q_1}),\dots,f(x_{q_v})$}. These values are then compared to the correct output \add{$y_{q_1}, \dots, y_{q_v}$} from the training data using a loss function. \add{The loss function is used to calculate the inaccuracy of predictions.} In the next step, the weights of the network are adjusted according to their influence on the loss function to reduce the loss function value in the next iteration (gradient descent). Once all \add{training examples} of the training set have been processed, the first \emph{epoch} of the training is completed. The training can be continued for several epochs until no further improvement of the error is observed.  We again refer to~\cite{goodfellow2016} for more details regarding the learning process.

\subsection{\add{Search Strategies}}
\add{The overall order in which nodes are explored is determined by a high-level search strategy.
We evaluate several well-known strategies, namely DFS, LDS, and BFS, which each depend on the branching and bounding networks to varying degrees. While DFS explores nodes with respect to their depth (deeper nodes are explored first), LDS explores nodes depending on the number of deviations from the search path recommended by the branching network. WBS is a best first search in which nodes with the lowest lower bound are explored first. In the following, we discuss each of the strategies in detail.} 

\subsubsection{Depth first search}

Algorithm~\ref{alg:dfs} shows the depth-first DLTS approach. The algorithm is called with the start state $s_0$ of the instance, stored in a node $n$, that has several properties. These are whether the associated (partial-) solution is complete, $\scomplete(n)$, the current cost of the associated (partial-) solution, $\scost(n)$, the depth of the node in the tree, $\sdepth(n)$, and the successor node\add{s} (children) of the current node, $\ssucc(n)$. Furthermore, the bounding network query frequency $k$, the lower bound uncertainty adjustment $d$, the branch pruning adjustment parameter $p$. \add{Additionally, the upper bound $ub$ (a global value), representing the cost of the best solution found so far, is set to its initial value $\ub_{ini}$.}

The function starts by checking whether the current node represents a complete solution and if the cost is better than the current best known cost. If this is the case, the upper bound is updated and the node is returned. On line~\ref{line:hlb} we compute a heuristic lower bound \add{($hlb$)} of the current node, but only at depths mod $k$, as querying the bounding network is expensive. We multiply the bounding network estimate by a factor $d$, between 0 and 1, to reduce the risk of overestimating the lower bound. This results in a weakened heuristic lower bound, but less likelihood of cutting off the optimal solution. Should the cost of the current node or the heuristic lower bound exceed the current upper bound, we return the empty set and define $\scost(\emptyset) := \infty$.

The branching network is queried on line~\ref{line:policy} for each successor node. \dnnp{} \add{is used to} form a probability distribution over all valid branches \add{and returns the probability of the child node $n'$ of $n$.} \add{The maximum probability value of the distribution is stored in $r$.} We can interpret \add{each} probability value as the confidence the network has that a particular successor node is the optimal successor for $n$. \add{We exclude any successors that are below a minimum probability threshold on line~\ref{line:mp}, which can be computed through one of several functions $\mprob$} that we describe in Section~\ref{sec:mp}. The list of successors is sorted by the prediction from the DNN, and nodes with a higher value are explored first.

\begin{algorithm}[t]
    \textbf{Input:} \add{A node $n$ of the search tree; bounding network query frequency $k$;
    lower bound uncertainty adjustment $d$; branch pruning adjustment parameter~$p$.}\\
    \textbf{Global:} \add{Cost of the best seen complete solution $\ub$ (Initial value: $\ub_{ini}$).}\\
    \textbf{Output:} \add{Node representing the best complete solution found (with costs below $ub$); $\emptyset$ otherwise.} \\
    \begin{algorithmic}[1]
        \Function{DLTS-DFS}{$n, k, d, p$}
            \If{$\scomplete(n)$ and $\scost(n) < \ub$}
                \State $\ub \gets \scost(n$\add{)}
                \ReturnNNL $n$
            \EndIf
            \State $\lb \gets -\infty$
            \If{$\sdepth(n) \bmod k = 0$}
                $\lb{} \gets \scost(n) + \dnnv{}(n) \cdot d$ \label{line:hlb}
            \EndIf
            \If{$\scost{}(n) \geq \ub$ or $\lb \geq \ub$ or CPU time exceeded}
                \Return $\emptyset$
            \EndIf
            \State $r \gets \max_{n' \in \ssucc(n)}\{\dnnp{}(n, n')\}$ \label{line:policy}
            \State $B \gets \{n' \in \ssucc(n) \stt \dnnp{}(n,n') \geq \add{\mprob{}(p, r, \scost(n), \ub)\}}$ \label{line:mp}
            \State Sort $B$ by $\dnnp{}(n, n')$ for each $n' \in B$, descending
            \ReturnNNL $\argmin_{n' \in B} \{\scost(\textsc{DLTS-DFS}(n', k, d, p)\}$
        \EndFunction
    \end{algorithmic}
    \caption{Depth first search based deep learning assisted heuristic tree search.\label{alg:dfs}}
\end{algorithm}

\subsubsection{Limited discrepancy search}

In our DFS, we order the search such that we always search nodes in the order recommended by the branching network. As with any heuristic, the branching network will sometimes be wrong. If a branching mistake happens near the root of the search tree, DFS will waste time searching entire sub-optimal subtrees before moving on to more promising areas. \add{LDS} addresses this by changing the search order so that the search proceeds iteratively by the number of \emph{discrepancies}. A discrepancy is a deviation from the search path recommended by the heuristic. The intuition of the search strategy is that the branching direction will be correct \emph{most of the time}. Thus, we ought to first examine solutions using only the advice of the heuristic, followed by solutions that ignore the advice of the heuristic a single time, followed by solutions ignoring the advice two times, and so on. 

\begin{figure}[tb]
    \centering
    \begin{subfigure}[t]{0.26\textwidth}
        \includegraphics[width=\textwidth]{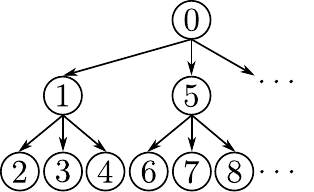}
        \caption{DFS path}
    \end{subfigure} \qquad
    \begin{subfigure}[t]{0.26\textwidth}
        \includegraphics[width=\textwidth]{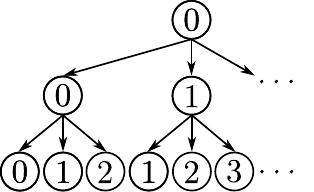}
        \caption{LDS discrepancies}
    \end{subfigure}
    \caption{Search ordering for DFS and LDS. \add{Branches are ordered from left to right as recommended by the branching network.}}
    \label{fig:ldsdfs}
\end{figure}

Figure~\ref{fig:ldsdfs} shows the search order for DFS and LDS in a typical tree search. Assume the branches in the figure are ordered from left to right according to the advice of the branching heuristic, i.e., it suggests going left first. Notice how with DFS, after branching to the left at the top of the tree the entire subtree is explored before moving on to the next subtree. The node explored fourth \add{by DFS}, for example, is not recommended by the heuristic, however it is nonetheless explored before the 6th node in the search, which is recommended by the heuristic after a single \add{discrepancy} in the root node.
DFS often examines nodes that have, according to the branching heuristic, a low probability of success before nodes with a high probability. LDS, however, searches in order of the likelihood of finding an optimal solution, according to the heuristic.

LDS is traditionally presented on binary search problems, i.e., each non-leaf node has exactly two child nodes. However, the branching factor in DLTS is not restricted to two. We adopt the generalized LDS scheme of~\cite{furcy2005}, in which child nodes are ordered, and the discrepancy is computed as the number of nodes away from the recommended node. \cite{furcy2005} also use a hash table to prevent cycles, however we do not implement this.

LDS is often implemented in an iterative process, in which a DFS explores all nodes of discrepancy 0, followed by a new DFS exploring nodes with discrepancy 1, then 2, and so on (see~\cite{korf1996}). While the method of~\cite{korf1996} avoids visiting any leaf node more than once, internal tree nodes can be visited multiple times. Querying the branching and bounding networks in our tree search is computationally expensive, so we do not want to repeat this work. We therefore use a priority queue approach instead of an iterative DFS, as shown in Algorithm~\ref{alg:lds}.

The algorithm accepts the parameters $n, \add{k, d, p, \ub_{ini}}$, which \add{are} the same as in the DFS. The algorithm starts by initializing the best known solution to nothing ($\emptyset$) and creates a priority queue with the root node solution. The algorithm then loops until the queue is empty, popping the node with the lowest discrepancy. We note that if \add{$\sdepth(n) < z$}, then we set the discrepancy of the node to 0 for the purposes of the queue \add{(with $z$ being a tunable parameter)}. This allows the search to open more nodes at the top of the tree before applying LDS. Ties between nodes of equal discrepancy are broken by examining nodes of higher depth first (i.e., those nodes closest to a leaf node) as in~\cite{sellmann2002}. If the popped node is the best seen so far, $\bestm{}$ is updated. Then, the bounding network is queried (depending on the value of $k$) and the child nodes are pruned if it is determined that they would be too expensive. Otherwise, branching is performed, using only those branches allowed by the branching network as in the DFS.

LDS is designed for trees in which there are only two branches. In the CPMP, and indeed in many optimization problems, the \add{number of branches} can be much higher \add{(in case of the CPMP, the number of branches is $S*(S-1)$, where $S$ is the number of stacks)}. This could pose an issue, \add{because only the order of the scores is taken into account and not their absolute value. In cases where two nodes are assigned a similar score, there is no reason to enforce an order between them. In these cases, we wish to assign both nodes the same discrepancy value.} We thus introduce a\add{n optional binning mechanism that reassigns} the discrepancies of nodes in $Q$ as follows. We first calculate the size of each bin by dividing the maximum probability output from the branching network by $b$, the number of bins. That is, each bin represents a probability range $[l_i, u_i)$, with $l_i = u_{i+1}$ for $i < b$ (i.e., the bins are sorted with higher probabilities first). For each potential branch of a node, we assign it a discrepancy according to the number of the bin it falls in to. 

\begin{algorithm}[t]
    \textbf{Input:} \add{Node $n$ representing the start state $s_0$; bounding network query frequency $k$;
    lower bound uncertainty adjustment $d$; branch pruning adjustment parameter $p$; initial value for $\ub$ named $\ub_{ini}$.} \\
    \textbf{Output:} \add{Node that can be reached from $n$ representing a complete solution with costs below $ub$; $\emptyset$ otherwise.} \\
    \begin{algorithmic}[1]
        \Function{DLTS-LDS}{$n, \add{k, d}, p, \add{\ub_{ini}}$}
            \State $\add{\ub \gets \ub_{ini}}$
            \State $\bestm \add{\gets \emptyset}$
            \State $Q \gets \{n\}$ \Comment{$Q$ is sorted by the node's discrepancy}
            \While{$Q$ is not empty and CPU time not exceeded}
                \State $n \gets \textsc{Pop}(Q)$
                \If{$\scomplete(n)$ and $\scost(n) < \scost(\bestm)$}
                    \State $\bestm \gets n$; \textbf{continue}
                \EndIf
                \State $\lb \gets -\infty$
                \If{$\sdepth(n) \bmod k = 0$}
                    $\lb{} \gets \scost(n) + \dnnv{}(n) \cdot d$ \label{line:hlblds}
                \EndIf
                \If{$\scost(n) < \scost(\bestm)$ and $\lb < \scost(\bestm)$}
                
                    \State $r \gets \max_{n' \in \ssucc(n)}\{\dnnp{}(n, n')\}$ 
                    \State $B \gets \{n' \in \ssucc(n) \stt \dnnp{}(n,n') \geq \mprob{}(p, r,\add{\scost(n)}, \scost(\bestm))\}$
                    \State $Q \gets Q \cup B$
                \EndIf
            \EndWhile
            \ReturnNNL \best
        \EndFunction
    \end{algorithmic}
    \caption{Limited discrepancy search based deep learning assisted heuristic tree search.\label{alg:lds}}
\end{algorithm}

\subsubsection{Weighted beam search}

A further alternative to DFS and LDS is WBS, which is a heuristic based on best first search. In best first search, nodes with the lowest lower bound are explored first and a heuristic version \add{of it}, beam search~\citep{russell2011}, has been widely applied in the AI and OR communities. Beam search limits the number of child nodes that are explored at each search node to a constant amount, $\beta$, \add{called the beam width}. This is the same concept as we use in our DFS and LDS, except we use \add{an adaptive} width.

For WBS, we compute the bound used for sorting the nodes of the search using a weighted sum of the cost of the node \add{$n$} plus the estimated lower bound of the node: \add{$wlb(n) = \alpha \cdot \scost(n) + \gamma \cdot \dnnv{}(n)$, with $\alpha$ and $\gamma$ being tunable parameters.} WBS is thus able to place more emphasis on the bounding network's prediction if desired \add{and} the branching network is only used to determine the beam width \add{using the function $\mprob$}. \add{Note that the bound for the pruning of the search tree is still computed as in DFS and LDS.} 

Since the pseudocode for \textsc{DLTS-WBS} is very similar to \textsc{DLTS-LDS}, we do not provide a separate code listing. The priority queue, $Q$, in Algorithm~\ref{alg:lds} is adjusted so that the sorting criterion is the heuristic lower bound as described. Furthermore, instead of only computing the lower bound when \add{$\sdepth(n) \bmod k = 0$}, it is computed for every node.

\subsection{Branch pruning functions}
\label{sec:mp}

\add{Using the function $\mprob$, we artificially limit the branches that are explored in a given node $n$. Only branches with a probability value above the value returned by the function $\mprob$ are explored.} \add{Often, this is only the case for a single branch of a node.} We define three simple functions, two of which adjust the \add{number of child nodes to be explored to the costs of the solution associated with $n$}. The intuition for this is that at the top of the tree \add{(where costs of the associated solutions are low)} picking the wrong branch can be extremely costly, as the optimal solution or near\add{-}optimal solutions may be removed from the tree. Mistakes further down in the search tree are not as bad, as the branching DNN will likely choose a good search path in a neighboring node.

All three functions accept the parameters $(p, r, \costc, \costb)$, which are the branch pruning adjustment parameter, the maximum probability assigned to any branch, \add{the cost $\scost(n)$ of the solution associated with $n$, and the cost of the best seen solution.} Each of the three $\mprob$ variants returns a value less than or equal to 1, and any branch assigned a probability by the branching DNN less than the value is pruned.

The function $\mprob\textsc{-Constant}$ is the simplest of all the functions, as it simply returns $p$ scaled to the largest probability $r$ and ignores all other input as follows:
\begin{equation}
    \mprob\textsc{-Constant} := r(1 - p).
\end{equation}

The constant version of $\mprob$ tends to be very expensive since the same number of branches are available at the top of the tree as at the bottom. The function $\mprob\textsc{-Quadratic}$ aims to decrease the \add{number of child nodes that are explored} more quickly so more areas of the tree can be searched within the time limit.
\begin{equation}
    \mprob\textsc{-Quadratic} := r\left(1 - p \frac{(\costb{} - \costc{})^2}{\costb{}^2}\right).
\end{equation}
Finally, we also introduce a log-based function as an alternative to the quadratic one:
\begin{equation}
\label{equ:log}
    \mprob\textsc{-Log} := r\left(1 - p(-\log(\frac{\costc{}}{ \costb{}})\right)).
\end{equation}

\section{Solving the Container Pre-marshalling Problem with Deep Learning Heuristic Tree Search}
\label{sec:cpmp_dlts}

DLTS is able to generate solutions for the CPMP without relying on branching and bounding heuristics or features designed by domain experts. However, a way to insert these (partial-) solutions into the DNNs \add{is} needed. In this section we first describe \add{the architecture of the branching and bounding DNNs for the CPMP. Additionally}, we describe which additional problem knowledge of the CPMP is provided to DLTS.

\subsection{DNN models}

\begin{figure}
    \centering
    \includegraphics[width=0.9\textwidth]{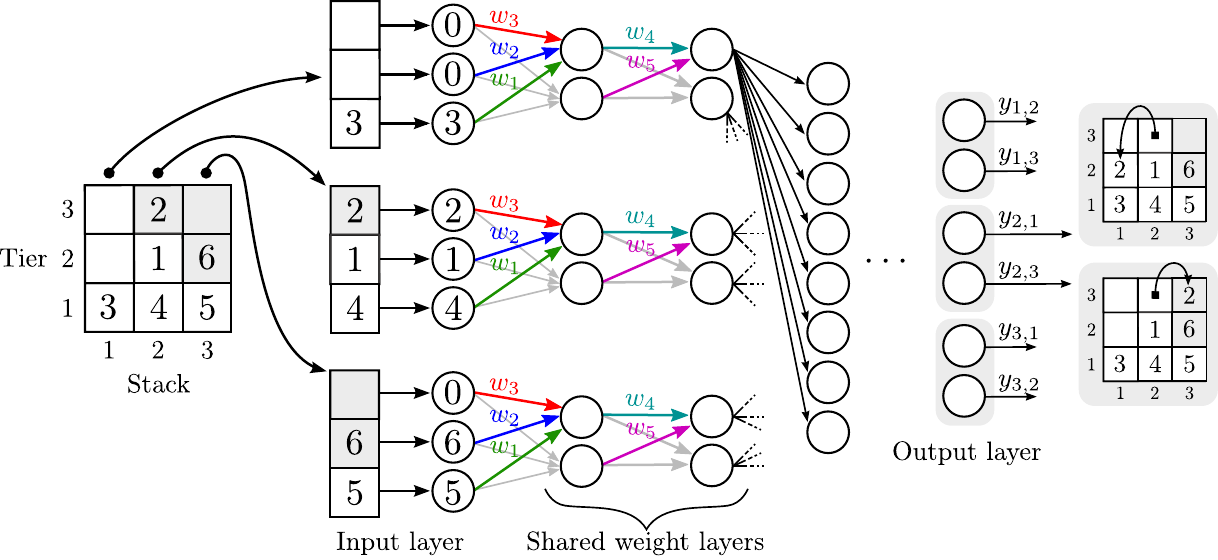}
    \caption{Branching DNN for the CPMP. \add{The shared weight layers are connected to the output layer by fully connected layers.}}
    \label{fig:cpmpdl}
\end{figure}

Figure~\ref{fig:cpmpdl} shows the structure of the branching DNN for the CPMP. The network is dependent on the size of the instance, however once trained for a particular instance size, instances with less stacks and tiers can also be solved by using dummy containers. The branching DNN's input layer consists of a single node for each stack/tier position in the instance. Directly following the input layer are \emph{locally connected} layers (as opposed to \emph{fully connected layers}) that bind each stack together. This provides the network with knowledge about the stack structure of the CPMP. We include several locally connected layers, followed by fully connected layers that then connect to the output layer.

We use a technique called \emph{weight sharing} directly following the input layer in which each tier is assigned a single weight, $w_i$, as opposed to assigning each container a weight. As can be seen in the figure, for example in the topmost tier, the weight $w_3$ is applied to each stack at that tier. The group value is multiplied by this weight, and then inserted into the next layer of the DNN. The propagation of the group values through these first layers can be thought of as a feature extraction process, where the same features are generated for each stack. The subsequent layers process these features and are fully connected: Each node processes its inputs with an activation function and sends its output into all nodes of the next layer. All nodes of the hidden layers use the rectifier activation function, defined as \add{$\textit{ReLU}(x) = \max\{0, x\}$}. 

In our experiments weight sharing leads to a slightly improved performance of the DNNs. However, it does not enable the DNNs to understand the symmetric nature of the CPMP, e.g.\add{,} that the minimum number of moves needed to solve an instance is independent of the order of the stacks. The DNN architecture we use could also be used for variations of the CPMP where the order of the stacks is of relevance, e.g.\add{,} when considering the time to move a container between stacks.

The output layer of the branching DNN consists of a node for each possible movement of a container from one stack to another stack \add{(including infeasible movements)}. \add{The DNN output can be understood as a probability distribution over these moves} with higher values corresponding to moves the DNN ``thinks'' are likely to lead to an optimal solution. These output values also provide a level of confidence, with higher values for a particular move meaning that the network is more certain about it being good.

The bounding DNN differs from the branching DNN only in terms of its output layer. There is only a single output node. \add{The training of the branching and bounding network is described in Section~\ref{sec:training}}.

\subsection{Additional problem knowledge}
\label{sec:knowledge}
The branching DNN can potentially select a move that is not feasible, for example moving a container to a stack that is already full. We filter such moves from the output of the DNN (corresponding to a simple domain specific heuristic), leaving only feasible moves. Furthermore, we do not allow moves that undo the directly preceding move. The work of \cite{idacpmp} and \cite{tanaka2017} point out that the CPMP can be solved significantly faster when avoiding symmetries by implementing specialized branching rules. We purposefully do not model these \add{ or any other advanced branching rules. This also means that the search can cycle in moves (for cycles of length $\geq$ 3). However, extensive cycling should be prevented by the pruning from the value network.}

\section{Computational Results}
\label{sec:compres}

We now evaluate DLTS on the CPMP. In our experiments, we attempt to answer the following questions:
\begin{enumerate}
    \item What effect do different DNN structures have on the performance of DLTS?
    \item What effect do different search strategies have on the performance of DLTS?
    \item Is DLTS competitive with state-of-the-art metaheuristics?
\end{enumerate}

To ensure a fair comparison of DNN structures and search strategies in research questions one and two, we use algorithm configuration either through a grid search or the configurator GGA~\citep{gga} to find high quality parameters for DLTS. With respect to research question three, we experiment on a variety of CPMP instances that we describe below. \add{Since DLTS requires more instances for training than are available in the literature, we generate instances that extend and generalize the instances from~\cite{caserta:voss:08}. However,} we also test DLTS on instances from the literature to show that the high quality performance of DLTS is not due to carefully selected instances. We also compare DLTS to the biased random-key genetic algorithm proposed in~\cite{brkgacpmp} and the target-guided heuristic from~\cite{WaJiLi15}. To our surprise, DLTS outperforms both of these heuristics despite having to learn the vast majority of its heuristic guidance \add{by} itself.

\subsection{Experimental setup}

Training DLTS requires a large number of instances. In total, we generate more than 900,000 instances of various sizes using the generator from~\cite{TiMa15} to train several DLTS instantiations. To ensure the applicability of DLTS to different types of CPMP instances, we create three different classes of instances: G1, G2 and G3. In G1, the group of every container is unique, as in the instances from~\cite{caserta:voss:09Coll}. In G2, every group is assigned to two containers. In G3, each group is assigned to three containers. We then make instances in each class in three different sizes defined as $S$x$T$ (stacks x tiers): 5x7, 7x7 and 10x7. We leave the two top tiers free so there is room to move containers around during pre-marshalling. We chose these sizes based on the sizes of real-world pre-marshalling problems in container terminals, which generally are no more than 10 stacks wide due to the maximum width of the cranes that move the containers, and are around 7 containers high due to safety restrictions.

We focus the training of DLTS on two versions of the above instance classes: G1 and G123, which is a combination of G1, G2 and G3. For each size (5x7, 7x7 and 10x7) we generate \add{a training dataset of 120,000 instances of G1 and a training dataset of 120,000 instances of G123, consisting of 40,000 instances each of G1, G2, and G3.} \add{We train different branching and bounding networks on each of these six datasets}. With G1, we test how well DLTS can adapt to a single type of instance. In testing G123, we determine whether or not DLTS can learn how to solve problems with a mixture of different instance types.

The branching and bounding networks are trained on reference solutions generated by the TT algorithm~\citep{tanaka2017}\add{, which performs an iterative deepening branch-and-bound search}. We attempt to solve all instances using TT with a time limit of 10, 20 and 30 minutes for 5x7, 7x7, and 10x7, respectively. If TT is unable to find an optimal solution within the time limit, the best solution found is used instead. \add{We further generate test sets for each instance size (5x7, 7x7 and 10x7) and instance class (G1, G2, and G3) consisting of 250 instances each and additional test sets containing 750 instances each of all instance classes (i.e., G123 instances) to test the DLTS approach as a whole.} We run TT on these instances for seven days and use the results for investigating the gap of DLTS to optimality.

We implement DLTS in Python 3 using \textit{keras 1.1.0}~\citep{chollet2015keras} with \textit{theano 0.8.2}~\citep{theano}  as the backend for the implementation of the DNNs.
All experiments are conducted using the Arminius Cluster of the Paderborn Center For Parallel Computing (PC$^2$) on Intel Xeon X5650 CPUs (2.67 GHz). All DNNs are trained on a single CPU using all six cores, resulting in a training time ranging from several hours to a few days. We run DLTS and evaluate the branching and bounding networks using a single thread. This means that, once trained, DLTS can be run on a typical desktop computer. This makes it especially useful for industrial applications.
 
\subsection{Experimental question 1: DNN configurations}

Configuring a DNN correctly is critical for it to perform well, and is a difficult problem in and of itself~\citep{domhan2015}. We therefore suggest three different possible configurations for the CPMP in which we adjust the number of shared weight layers (SWL) and non-shared weight layers (NSWL) for both the branching and bounding networks. All networks are trained using the Adam optimizer~\citep{kingma2014adam}, which is based on a gradient descent.
\add{We first explore the performance of different branching networks (in DLTS) in Section~\ref{sec:branching_networks}. We train branching networks of different sizes on the G123 datasets and evaluate the prediction quality of each network on 30,000 validation instances of the same class/size as the training dataset. We then insert each of the branching networks into DLTS to evaluate their performance on additionally generated validation datasets (each consisting of 300 G123 instances). The DFS search strategy and the log-based branch pruning function (shown in Equation~\ref{equ:log}) are used with a $p$ value found through a grid search. In Section~\ref{sec:bounding_networks}, we evaluate the performance of several branching networks. Training and evaluation of the bounding networks is done similarly to the branching networks. To evaluate the performance impact of the bounding network on the search for CPMP solutions, we use each bounding network together with the best performing branching network from Section~\ref{sec:branching_networks} in DLTS. We use the DFS search strategy with the log-based branch pruning function and tune the parameters $p$, $d$, and $k$ through a grid search\footnote{In later experimental questions, we use an algorithm configurator~\citep{gga} to set parameters, but avoid it on these first experimental questions due to the high computational cost.}.}

\subsubsection{Branching networks}
\label{sec:branching_networks}

Table~\ref{tab:policy_networks} shows the validation performance of the branching networks on G123. The learning rate for the Adam optimizer was set to 0.001 (the default value) for all networks, except for those trained on the 10x7 instances. For these, we set the learning rate to 0.0005 to \add{delay} overfitting. Higher rates represent more aggressive adjustments of the DNN weights. We use the \emph{early stopping} termination criteria, which stops the training after no performance improvement on the validation set is seen for a predetermined number of epochs (in our case 50). 

The columns of the table are as follows. The number of shared \add{weight} layers (\add{SWL}) and non-shared \add{weight} layers \add{(NSWL)} are given. The number of weights is the number of arcs in the DNN between perceptrons. We use the loss function \emph{categorical crossentropy} (CCE) to judge the performance of the DNN. CCE measures the distance of the output of the DNN to the desired probability distribution \add{(shown in Equation \ref{kss})}. A key advantage of CCE over the classification error is that it not only penalizes incorrect predictions, but also correct predictions that are weak. For example, a DNN suggesting a correct move with only slightly higher confidence than incorrect moves will receive a worse CCE value than a DNN that assigns a high confidence value to the correct move. The accuracy refers to the percentage of the validation set for which the DNN predicts the correct move. 

\add{For DLTS we provide the average relative gap to optimality for each dataset computed as}
\begin{equation}
\add{gap =
\frac{\text{Total number of DLTS moves on the dataset}}{\text{Total number of optimal moves on the dataset}} - 1.}
\label{eq:gap}
\end{equation}
We also provide the average time to solve the validation instances. A positive insight from these results are that lower CCE values also correspond to lower gaps. \add{This indicates that the CCE is a suitable loss function for the training of the branching networks. Note, that the shown average values do not hide a few really bad executions of the algorithm. The maximum gap of all individual solutions (2700 in total) generated for the branching network evaluation is 18.4\%.} A second insight is that bigger networks are not always better. For example, for 10x7 the network with 471,729 weights outperforms the network twice its size in terms of CCE, accuracy and DLTS gap. It is clear, however, that having a network that is too small hampers learning, especially on large instances. Since the predictions of small networks can \add{usually} be computed faster, it would be reasonable to expect them to have an advantage over large networks. However, networks that are \emph{too} small sacrifice too much predictive accuracy, as seen for all three instance sizes.

\begin{table}[tb]
\footnotesize
\centering
\caption{Validation performance of different branching networks trained on the G123 dataset.}
\label{tab:policy_networks}
\begin{tabular}{l|rrr|rr|rr}
               & \multicolumn{3}{c|}{Network Properties}         & \multicolumn{2}{c|}{Validation} & \multicolumn{2}{c}{DLTS} \\ \hline
Size & SWL & NSWL & Weights & CCE        & Accuracy       & Gap (\%)            & Time (s)            \\ \hline 
5x7            &        2         &            3         &   63,923 &          0.563             &      80.18     &   1.53     &   39.51             \\ 
               &        3         &            3         &  118,089 &          0.532             &       81.27    & 1.27          &     37.01        \\ 
               &       3          &            4         & 214,591  &        0.538               &       81.29    & 1.31                  &         34.85             \\ 
\hline 
7x7            &       2          &            3         & 125,629  &        0.740                &       75.74    &     2.77         &       36.02          \\ 
               &       3          &            3         & 230,433  &       0.693                &       77.12    & 2.14          &    55.17          \\ 
               &        3         &            4         & 417,599  &      0.713                 &       76.58    & 2.32      &    55.97        \\ 
\hline 
10x7           &        2          &           2         &  259,363 &         0.926              &      69.81                            &      4.20             &       57.90               \\ 
               &        2          &          3          & 471,729  &      0.839                    &     72.15           &      3.01             &       57.05             \\ 
               &        3          &          4          &  851,486 &      0.894                  &      70.55          &      3.60           &    57.79    
\end{tabular}
\end{table}

\begin{figure}[b]
    \begin{center}
        \begin{subfigure}[b]{0.3\textwidth}
            \includegraphics[width=\textwidth]{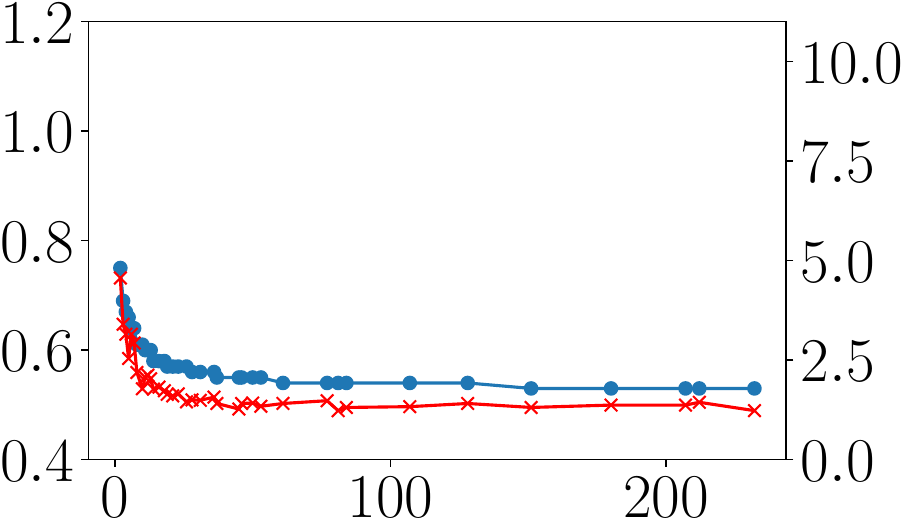}
            \caption{5x7} 
            \label{fig:valperf5x7}
        \end{subfigure} \;
        \begin{subfigure}[b]{0.3\textwidth}
            \includegraphics[width=\textwidth]{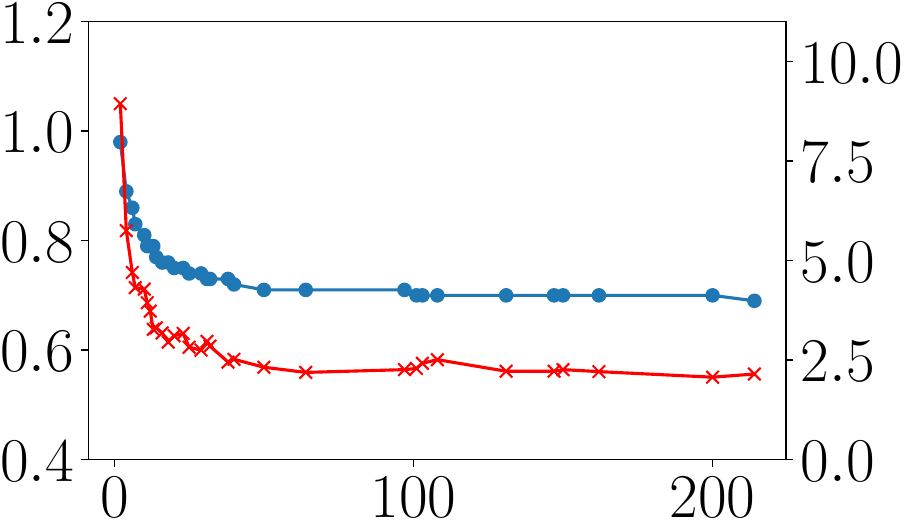}
            \caption{7x7}
            \label{fig:valperf7x7}
        \end{subfigure} \;
        \begin{subfigure}[b]{0.3\textwidth}
            \includegraphics[width=\textwidth]{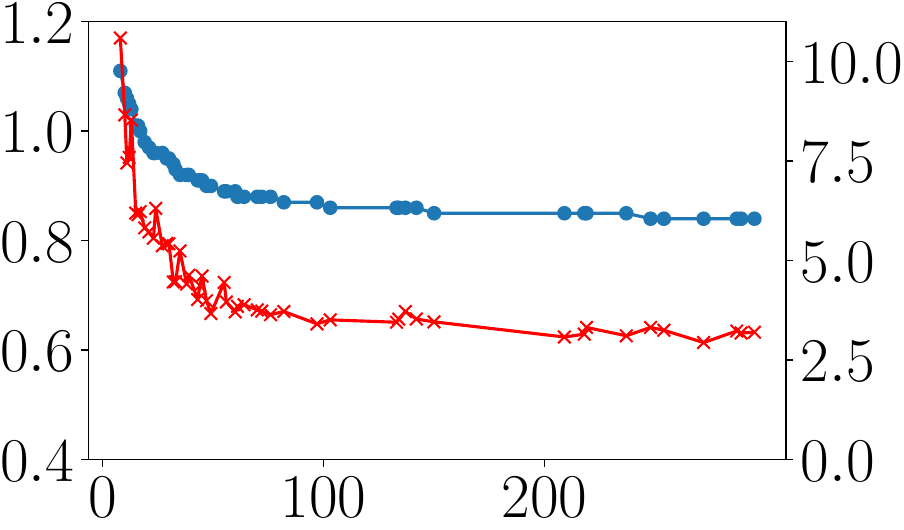}
            \caption{10x7}
            \label{fig:valperf10x7}
        \end{subfigure}
    \end{center}
    \caption{Validation CCE of the branching network (blue circles) \add{on the left axis} and DLTS gap in percent (red crosses) \add{on the right axis for} each \emph{epoch} of the DNN training.}
    \label{fig:valperf}
\end{figure}

\add{Figure~\ref{fig:valperf} shows the performance of the ``medium'' sized branching networks over the course of their training. Each time that a new best validation CCE is observed, we insert the corresponding network into DLTS and search for solutions to the validation set instances. In case that a solution is found for all instances we include the observed gap and the validation CCE of the network in the figure.} 
Examining the training for 7x7 instances, we note that DLTS achieves a gap of around 9\% to the best solutions found even for a DNN that has a CCE value of 0.95 -- meaning it makes many mistakes. A gap of 9\% is already better than heuristics from the literature for the CPMP on instances of this size, such as the corridor method~\citep{caserta:voss:09Coll} or LPFH~\citep{Exposito}.

\subsubsection{Bounding networks}
\label{sec:bounding_networks}

Table~\ref{tab:valnetperf} shows the validation performance for the bounding networks. Since the bounding DNN is performing a regression, we swap CCE for the mean squared error (MSE) and provide the mean absolute error (MAE) instead of the accuracy. As in the case of the branching networks using the CCE criterion, the MSE score correlates with the DLTS gap we find. However, in contrast to the case of the branching networks, the larger networks result in nearly the same performance as the ``medium'' sized networks.

\begin{table}[tb]
\footnotesize
\centering
\caption{Validation performance of different bounding networks trained on the G123 dataset.}
\label{tab:valnetperf}
\begin{tabular}{l|rrr|rr|rr}
               & \multicolumn{3}{c|}{Network Properties}         & \multicolumn{2}{c|}{Validation} & \multicolumn{2}{c}{DLTS} \\ \hline
Size & SWL     & NSWL     & Weights & MSE        & MAE       & Gap (\%)            & Time (s)            \\ \hline
5x7            &        2         &            2         &  5,258  &       0.522               &      0.556    &                 1.15  &         41.61            \\ 
               &        3         &            3         &  21,057 &       0.300               &      0.390    &  0.95  &           25.57           \\ 
               &       3          &            4         &  44,255 &       0.279               &      0.376     &  0.94 &           26.09           \\ 
\hline 
7x7            &       2          &            2         &  10,018 &       0.475                &     0.499      &                1.98   &          44.06            \\ 
               &       3          &            3         &  39,999 &       0.307                &     0.372      & 1.68   &          22.85            \\ 
               &        3         &            4         &  84,421 &       0.298                &     0.354      & 1.72   &          34.27            \\ 
\hline 
10x7           &        2          &           2           & 20,098   &      0.490                &   0.488     &                2.50   &          35.19            \\ 
               &       3           &           3           & 80,172   &      0.395                &   0.410     & 2.16    &          48.59            \\ 
               &       3           &           4           & 169,660  &      0.392                &   0.399     & 2.16    &          47.37     
\end{tabular}
\end{table}

\subsection{Experimental question 2: Search strategy evaluation}

We now compare the three proposed search strategies across our datasets. To ensure a fair comparison between the strategies, we tune each search strategy with DLTS using GGA~\citep{gga} for a maximum of seven days. We give the tuning procedure the freedom to select the branching and bounding DNN (from those trained on G123), as well as to tune other DLTS parameters detailed in Section~\ref{sec:dlts}. \add{We evaluate the performance on additionally generated validation sets (with 250 instances each for G1, G2, and G3 and 750 instances for G123).} Table~\ref{tab:searchstrategycomp} provides the results in terms of the gap from the best known solution to each instance in the validation set (note that in nearly all cases this is an optimal solution). \add{A star indicates that not all instances were solved}, meaning that the value in the table cannot be used for a direct comparison between strategies.

\begin{table}[tb]
\footnotesize
\centering
\caption{Comparison of search strategies on the validation set for G1, G2 and G3 for all instance sizes \add{using branching and bounding networks trained on G123.}}
\label{tab:searchstrategycomp}
\begin{tabular}{rr|rrr|rrr}
     &      & \multicolumn{3}{c|}{Gap (\%)} & \multicolumn{3}{c}{Avg. Time (s)}  \\ \hline
  Size   & Class & DFS      & LDS      & WBS   & DFS      & LDS      & WBS         \\ \hline
5x7  & G1 &  2.03        &   \textbf{0.81}     &  *1.62      &     4.09    &  47.59              &     *59.62                         \\
     & G2 &   1.88       &   \textbf{0.53}     &   1.21     &     2.92    &  41.58             &    58.09                           \\
     & G3 &  1.67         &  \textbf{0.52}    &  *0.84      &    2.26    &   35.27            &   *57.24                           \\ 
     & G123 &  1.87         &  \textbf{0.62}  &    *1.23       &    3.09          &     41.48          &   *58.32                            \\ \hline
7x7  & G1 &    \textbf{2.14}       &    2.23        &   2.83     &   42.01           &    59.9     &  25.26                             \\
     & G2 &    \textbf{1.55}       &   1.65         &   2.21     &   38.38           &    59.68    & 22.29                              \\
     & G3 &    \textbf{1.23}       &   1.24         &   2.05     &   32.58           &    59.32    &  19.52                             \\ 
     & G123 &   \textbf{1.64}      &   1.71         &   2.37     &    37.66      &   59.63          &     22.36                          \\ \hline
10x7 & G1 &    \textbf{2.66}       &   3.16       &    *3.19      &    49.07      &    59.21         &  *59.90                             \\
     & G2 &   \textbf{2.16}        &   2.34       &   2.64         &    45.43      &    56.86         &   59.90                            \\ 
     & G3 &   \textbf{1.90}        &    2.12      &   2.26         &   42.28      &     55.93          &   59.90                            \\ 
     & G123 &  \textbf{2.24}  &   2.54      &   *2.70       &     45.59         &    57.33           &   *59.90                         
\end{tabular}
\end{table}

LDS and DFS provide the best overall performance as they find solutions to every instance they are given. While LDS provides under half the gap of DFS for 5x7 instances, we note that this usually means LDS finds solutions with roughly one less move than DFS. On larger instance sizes, DFS again outperforms LDS. Given that neither DFS or LDS dominates the other on all instance categories, it is not possible to draw any sweeping conclusions regarding the two search strategies. The main takeaway, however, is that it is important to use an algorithm configurator when creating a DLTS approach, since the performance of the search strategies varies.

\medskip

We note that the runtime of the results we obtain probably could be improved \add{by} using a faster programming language or using the GPU instead of the CPU for the neural networks. Table~\ref{tab:nnodes} shows the number of tree nodes we process during search compared to the TT method, which performs an iterative deepening branch-and-bound programmed in C. The number of nodes DLTS opens in comparison to TT is many orders of magnitude less, for a penalty of usually only one or two moves (a couple of percent) gap to optimality. For example, on the 10x7 instances, we explore roughly 36,000 nodes on average with LDS. The TT method explores upwards of 5 million nodes \emph{per second}. This is a clear indication that the \add{guidance} of the DNNs is extremely effective.

\begin{table}[tb]
\footnotesize
\centering
\caption{Number of nodes opened in the search tree during search on the validation set for TT and DLTS.}
\label{tab:nnodes}
\begin{tabular}{rr|rrr|rrr}
    &&  \multicolumn{3}{c|}{Avg. Opened Nodes (log)} & \multicolumn{3}{c}{Avg. Time (s)}  \\ \hline
  Size   & Class &  TT       & DLTS-DFS & DLTS-LDS   & TT         & DLTS-DFS & DLTS-LDS        \\ \hline
5x7  & G123 &   20.0     &    9.2       & 10.8    &  110.13 &  3.09 &   41.48                     \\
7x7   & G123  &  22.2       &  11.6  &    10.7                         & 875.03 & 37.66    &  59.63                    \\
10x7   & G123  &   24.6 & 11.5     & 10.5      & 9605.91 &  45.59      & 57.33                  \\
\end{tabular}
\end{table}

\subsection{Experimental question 3: Comparison to the state-of-the-art}

We compare DLTS to the state-of-the-art metaheuristic BRKGA from~\cite{brkgacpmp} in Table~\ref{tab:cvcomp}. We train DLTS on the G1 and on the G123 \add{training} datasets and report the performance of each on the test sets of G1, G2, G3, and G123. \add{We refer to the version trained on G1 as DLTS-G1 and to the version trained on G123 as DLTS-G123.} For DLTS-G123, we use the configuration with the best performance on the validation set in Table~\ref{tab:searchstrategycomp} for each instance size. For DLTS-G1, we configure a new set of parameters for each instance size on the G1 data, leaving the search strategy open, as well as all parameters tuned for DLTS-G123 \add{(including the selection of the networks)}. For a fair comparison to the state-of-the-art, we also tune the BRKGA algorithm \add{with GGA} on the G123 data.

While the BRKGA finds its best solution faster than DLTS for all instance sizes, the solutions it finds have optimality gaps between 3 and 23 times larger than DLTS. The importance of training DLTS on instances drawn from the same distribution as those it will see during testing is emphasized by the DLTS-G1 gaps. While DLTS-G1 sometimes does perform better than the BRKGA on data it was not trained for, such as for G2 on all sizes, as the instances become increasingly different (G3), performance suffers. DLTS-G123, however, shows high quality results for all instance groups, meaning that training across a wide range of different types of instances does not hurt performance.

\begin{table}[tb]
\footnotesize
\centering
\caption{Comparison to state-of-the-art metaheuristics on the test set.}
\label{tab:cvcomp}
\begin{tabular}{rr|rrr|rrr}
     &      & \multicolumn{3}{c|}{Gap (\%)} & \multicolumn{3}{c}{Avg. Time (s)}  \\ \hline
  Group   & Class & BRKGA      & DLTS-G1      & DLTS-G123   & BRKGA      & DLTS-G1      & DLTS-G123         \\ \hline
5x7  & G1 &   17.22  &   0.94     &  \textbf{0.75}     &    27.29          &    49.74           &            44.59                \\
     & G2 &   15.69    &   9.34     &   \textbf{0.66}  &    20.48          &    50.15           &           40.26                    \\
     & G3 &14.85      &   16.38    &    \textbf{0.63}        &  14.80        &  50.20           &  34.98      \\
     & G123 &  15.95     &   8.67         &   \textbf{0.68}         &     20.86        &     50.03          &  39.95
     \\ \hline
7x7  & G1 &   9.73        &    \textbf{1.64}    &  2.11      &   10.53   &     59.90          &           43.86                    \\
     & G2 &   9.13       &    7.42        &   \textbf{1.73}    & 10.03   &  59.90             &      38.65                         \\
     & G3 &   8.07        &   18.25         &   \textbf{1.34}   &  9.54    &  59.90             &       33.42     \\
     & G123 &  8.99         &    8.96        &    \textbf{1.73}        &     10.03    &     59.90         &     38.65                         \\ \hline
10x7 & G1 &   7.59      &   \textbf{2.65}         & 2.72 &    29.81      &   56.52            &           47.81                    \\
     & G2 &   7.11        &    5.65      &   \textbf{2.19}     & 29.54 &   57.53            &    41.43                           \\
     & G3 &   6.64        &   11.67         &    \textbf{2.06}  &  28.23     &  57.28             & 39.65     \\
     & G123 &   7.12        &  6.61      &    \textbf{2.33}        &   29.20           &    57.11          &     42.96                        
\end{tabular}
\end{table}

As a final test of DLTS, we solve instances from the CV dataset from~\cite{caserta:voss:09Coll} in Table~\ref{tab:soacv}. \add{Each instance group in the dataset consists of 40 instances with the same number of stacks $S$ and tiers $T$ (shown in Table~\ref{tab:soacv}).} We note that we perform no training or validation on these instances; we only run the DLTS approaches trained on instances generated to be similar to them. We report the average number of moves each solution procedure requires to \add{sort all stacks}, along with the average number of moves when solved to optimality. Unsurprisingly, DLTS-G1 outperforms DLTS-G123, since the CV instances have the same structure as G1: a single group per container. 

DLTS-G1 achieves the best gap to optimality to date, and in less than 60 seconds of run time. Averaging only 42.17 moves over the 40 instances of the CV 5-10 category, its solutions are usually only about 1 move away from optimal, whereas BRKGA and BS-B~\citep{WaJiLi15} are between 3 and 4 moves, respectively. In real container terminals, hundreds of CPMPs are solved for the various groups of stacks in the terminal, meaning improving the heuristic solution by even 2 moves could result in hundreds or even thousands of less pre-marshalling crane movements.

\setlength\tabcolsep{1.5pt}
\begin{table}[tb]
\footnotesize
\centering
\caption{Average number of moves for BS-B \add{as reported in \citep{WaEtal17}}, BRKGA and DLTS on the CV instances.}
\label{tab:soacv}
\begin{tabular}{lrr|rrrrr|rrrrr}
    &&&  \multicolumn{5}{c|}{Avg. Moves} & \multicolumn{4}{c}{Avg. Time (s)}  \\ \hline
  Group   &  $S$ & $T$    &  Opt. & BS-B & BRKGA & DLTS-G1      & DLTS-G123    & BS-B & BRKGA    & DLTS-G1      & DLTS-G123         \\ \hline
CV 3-5  & 5 & 5 &  10.15 &   10.45      &   \textbf{10.33}    & 10.35 & 10.40 &    0.01         &   1.19  & 1.06 &  1.03                   \\
CV 4-5   & 5 & 6 &  17.85 &  18.90   & 18.75      & \textbf{17.90} & 18.05 &         0.11          &   5.38         & 12.11 & 10.47         \\
CV 5-5   & 5 & 7 &  24.95 &  27.38      & 27.88    & \textbf{25.10} & \textbf{25.10} &     0.39          &  25.23          &  46.32 & 36.73 \\  \hline
CV 3-7   & 7 & 5 &  12.80 &  13.13     & 12.93     & \textbf{12.90} & 13.30 &          0.03        &   1.17         & 42.40 & 0.30 \\
CV 4-7   & 7 & 6 &  21.82 &  23.15   & 22.73      & \textbf{22.07} & 22.30 &        0.33          &   4.41         & 59.84 & 4.04    \\  
CV 5-7   & 7 & 7 &  31.48 &  34.20      & 33.83     & \textbf{31.98} & 32.08 &         1.51          &  20.77          & 59.91 &  42.26  \\ \hline
CV 5-10   & 10 & 7 &   41.23  & 44.85      & 44.00     & \textbf{42.17} &  42.23    &    7.46          &  14.53          & 54.97 &49.37                         \\
\end{tabular}
\end{table}

\section{Conclusion and future work}
\label{sec:concl}

We presented DLTS, a heuristic tree search that uses deep learning as a search guidance and pruning mechanism and applied it to a well-known problem from the container terminals literature, the container pre-marshalling problem. We showed that DLTS finds \add{better solutions than  state-of-the-art approaches} on real-world sized instances from the literature. DLTS does this with very little input from the user regarding \add{the} problem; it mostly relies on the provided (near-) optimal solutions to learn how to build a solution all on its own. To the best of our knowledge, DLTS is the first search approach for \add{an optimization problem} that allows a learned model to fully control decisions during search and is able to achieve state-of-the-art performance.

There are many avenues of future work for DLTS. One clear way forward is applying DLTS to other optimization problems, such as routing/scheduling problems. \add{DLTS is a promising approach for problems that
\begin{enumerate*}[label=\textit{\alph*)}]
\item allow a sequential solution construction (and for which construction heuristics have performed well in the past) and 
\item have a (partial-) solution and instance structure that allow for a quick evaluation of the DNNs. 
\end{enumerate*}
Other areas of future work include the usage of reinforcement learning as} in~\cite{silver2016} to further improve performance. Moreover, there are many changes to DLTS that can be made, such as reconfiguring the DNN or adjusting the search procedure, that may improve the performance in terms of runtime and solution quality.

\section*{Acknowledgment}{We thank Yuri Malitsky for insightful discussions about this work, and the Paderborn Center for Parallel Computation (PC$^2$) for the use of the Arminius cluster.}

\bibliography{bib}

\end{document}